\documentclass[runningheads]{llncs}

\usepackage{eccv}

\usepackage{eccvabbrv}
\usepackage{enumitem}
\usepackage{graphicx}
\usepackage{booktabs}
\usepackage{amsmath}
\usepackage{multirow}
\usepackage[accsupp]{axessibility}  
\usepackage[ruled,vlined]{algorithm2e}

\usepackage[pagebackref,breaklinks,colorlinks,citecolor=eccvblue]{hyperref}

\begin{document}

\title{Simba: Mamba augmented U-ShiftGCN for Skeletal    
Action Recognition in Videos} 

\titlerunning{Simba}

\author{Soumyabrata Chaudhuri\inst{1} 
Saumik Bhattacharya\inst{2}}

\authorrunning{Chaudhuri et al.}

\institute{Indian Institute of Technology, Bhubaneswar, Odisha, India
\email{chaudhurisoumyabrata@gmail.com}\\ \and
Indian Institute of Technology, Kharagpur, West Bengal, India\\
\email{saumik@ece.iitkgp.ac.in}}

\maketitle

\begin{abstract}
  Skeleton Action Recognition (SAR) involves identifying human actions using skeletal joint coordinates and their interconnections. While plain Transformers have been attempted for this task, they still fall short compared to the current leading methods, which are rooted in Graph Convolutional Networks (GCNs) due to the absence of structural priors. Recently, a novel selective state space model, Mamba, has surfaced as a compelling alternative to the attention mechanism in Transformers, offering efficient modeling of long sequences.
  In this work, to the utmost extent of our awareness, we present the first SAR framework incorporating Mamba. Each fundamental block of our model adopts a novel U-ShiftGCN architecture with Mamba as its core component. The encoder segment of the U-ShiftGCN is devised to extract spatial features from the skeletal data using downsampling vanilla Shift S-GCN blocks. These spatial features then undergo intermediate temporal modeling facilitated by the Mamba block before progressing to the encoder section, which comprises vanilla upsampling Shift S-GCN blocks. Additionally, a Shift T-GCN (ShiftTCN) temporal modeling unit is employed before the exit of each fundamental block to refine temporal representations. This particular integration of downsampling spatial, intermediate temporal, upsampling spatial, and ultimate temporal subunits yields promising results for skeleton action recognition. We dub the resulting model \textbf{Simba}, which attains state-of-the-art performance across three well-known benchmark skeleton action recognition datasets: NTU RGB+D, NTU RGB+D 120, and Northwestern-UCLA. Interestingly, U-ShiftGCN (Simba without Intermediate Mamba Block) by itself is capable of performing reasonably well and surpasses our baseline.
\end{abstract}

\section{Introduction}
\label{sec:intro}

Recognition of human actions based on skeletons has increasingly captivated attention due to its computational efficacy and ability to withstand fluctuations in the environment and differences in camera perspectives. A notable benefit of skeleton-based action recognition lies in the ease of acquiring body keypoints through sensors like Kinect\cite{6190806} or dependable pose estimation algorithms\cite{8099626}. Therefore, this makes pose a mode reliable modality in comparison to conventional RGB, optical flow or depth-based methods. 

Recently, Graph Convolution Networks (GCNs) \cite{kipf2016semi} have found widespread application in modeling non-Euclidean data. Yan et al. \cite{Yan_Xiong_Lin_2018} were among the first to conceptualize joints and their interconnections as nodes and edges within a graphical structure. They utilized a Graph Convolutional Network (GCN) on this predefined graph to analyze the interactions among joints. Since then, GCNs have become the dominant choice for skeleton-based action recognition tasks. Several variations of GCNs, encompassing multiple modalities (e.g., joint, bone, joint velocity, and bone velocity) \cite{cheng2020shiftgcn,cheng2021extremely,song2020stronger} or multi-view graph representations as seen in MV-IGNet \cite{wang2020learning}, have been utilized to address this challenge of capturing internode relationships. Furthermore, methods based on graph transformers like ST-TR \cite{plizzari2021spatial} and DSTA \cite{shi2020decoupled} have been explored for skeleton action recognition over time.

However recently, there has been a significant shift in paradigm for modeling long sequences. Mamba \cite{mamba}, a selective structured state space sequence model (S6) has demonstrated remarkable prowess in efficiently modeling lengthy temporal data in language domain and genomics. This naturally raises several questions: 1) \textit{Can Mamba demonstrate effectiveness in encapsulating graphical relationships?} 2) \textit{Can it be further leveraged to efficiently model the temporal sequence of graph snapshots from a video?}  In this study, to the best of our knowledge, we introduce the inaugural skeletal action recognition framework incorporating Mamba which operates on temporal graph data. Each constituent module within our novel Simba model incorporates a U-ShiftGCN architecture, wherein Mamba serves as its fundamental core. The encoder segment of the U-ShiftGCN is crafted to extract rich spatial features from skeleton data using a downsampling vanilla Shift S-GCN \cite{cheng2020shiftgcn} blocks. Subsequently, these spatial features undergo intermediate temporal modeling facilitated by the Mamba block before advancing to the encoder section, which consists of vanilla Upsampling Shift S-GCN blocks. Additionally, a Shift T-GCN (ShiftTCN) \cite{cheng2020shiftgcn} temporal modeling unit is deployed preceding the exit of each constituent block to enhance temporal representations. Interestingly, the U-ShiftGCN architecture has itself never been explored before and achieves better performance than our baseline. This particular fusion of downsampling spatial Shift S-GCN, intermediate temporal Mamba and upsampling spatial Shift S-GCN followed by ShiftTCN's final temporal aggregation yields promising outcomes for skeleton action recognition as authenticated by our results. Notably, our model achieves state-of-the-art performance across three renowned benchmark skeleton action recognition datasets: NTU RGB+D, NTU RGB+D 120, and Northwestern-UCLA.

Our main contributions can be summarized as follows:
\begin{itemize}[label=$\triangleright$]
  \item We propose, to the best extent of our knowledge, the first skeleton action recognition (SAR) framework to incorporate mamba for temporal sequence modeling on graph data. 
  \item Our model, Simba, surpasses the previous state-of-the-art for SAR task on three popular benchmark datasets. 
  \item Remarkably, the derivative of our Simba framework, U-ShiftGCN, stands as a novel exploration in its own right, showcasing its ability to exceed baseline performance.
\end{itemize}

The subsequent sections of this paper are structured as follows: In Section \ref{sec:related_works}, we delve into an examination of extant techniques pertinent to skeletal action recognition (SAR) and underscore the significance of mamba in facilitating prolonged sequence modeling in an efficient way. Our proposed methodology is expounded upon in Section \ref{sec: method}. Section \ref{sec:exp} presents the empirical findings concerning our proposed approach, complemented by an exhaustive juxtaposition against state-of-the-art (SOTA) methodologies. Ultimately, the discourse is drawn to a close in Section \ref{sec: conclusion}, where we deliberate upon the salient observations and delineate future avenues of exploration within the ambit of our proposed research.

\section{Related Works}
\label{sec:related_works}

In this section, we delineate the primary literature pertinent to our research concerning skeleton representation, alongside the works making progress in the direction of efficiently modeling long sequence data.

\subsection{Skeleton-based action recognition}
In previous years, Recurrent Neural Networks (RNNs) \cite{article,Du2015HierarchicalRN} and Convolutional Neural Networks (CNNs) \cite{2017PatRe..68..346L,ke2017new} were commonly employed for the task of skeleton-based human action recognition. Nonetheless, these methods tend to overlook the spatial interactions among joints. Consequently, the prominence of Graph Convolutional Networks (GCNs) in this field has increased, as they adeptly capture spatial configurations through graph modeling.

\textbf{GCN-based approaches} The advent of GCN-based methodologies was spearheaded by Yan et al. \cite{Yan_Xiong_Lin_2018}, who initially utilized GCNs \cite{kipf2016semi} to capture joint correlations and underscored their effectiveness in action recognition. Subsequently, various adaptations of GCN, such as ShiftGCN \cite{cheng2020shiftgcn} and ShiftGCN++ \cite{cheng2021extremely}, were introduced to address the challenges in Skeleton Action Recognition (SAR). Instead of relying on cumbersome regular graph convolutions, ShiftGCN integrates novel shift graph operations and lightweight point-wise convolutions. These operations offer flexible receptive fields for both spatial and temporal graphs. Its successor, ShiftGCN++, is an exceptionally computationally efficient model specifically designed for low-power and low-cost devices with constrained computing capabilities.

\textbf{Transformer-based approaches} Recently, Transformer-based strategies have emerged as an alternative approach to tackling this challenge, with a primary focus on managing the additional temporal dimension. For instance, \cite{plizzari2021spatial} proposed a dual-stream model incorporating spatial and temporal Self-Attention mechanisms to capture intra- and inter-frame correlations, respectively. In contrast, DSTA-Net \cite{shi2020decoupled} employs a Transformer that alternates between modeling spatial and temporal dimensions. However, these approaches have not achieved comparable efficacy with state-of-the-art GCN-based methodologies . This lack of performance parity is attributed to their adherence to conventional Transformer designs, which fail to account for the unique characteristics of skeleton data.

\subsection{Long Sequence Modeling}
The effectiveness of self-attention is attributed to its dense information routing within a context window, enabling it to model intricate data patterns. However, this property entails inherent limitations: an incapacity to model beyond a finite window and a quadratic increase in computational complexity relative to window length. Recently, \textbf{structured state space sequence models (SSMs)} \cite{gu2021combining,gu2021efficiently} have emerged as a promising architectural class for sequence modeling. These models amalgamate elements of recurrent neural networks (RNNs) and convolutional neural networks (CNNs), drawing inspiration from classical state space models \cite{kalman1960new}. They exhibit remarkable computational efficiency, with linear or near-linear scaling in sequence length, and can be implemented either as recurrent or convolutional operations. SSMs function as standalone sequence transformations that can seamlessly integrate into end-to-end neural network architectures.

\textbf{H3} \cite{fu2022hungry} extended this recurrence using S4, presenting an architecture featuring an SSM flanked by two gated connections. Additionally, H3 introduced a standard local convolution, termed as a shift-SSM, preceding the primary SSM layer. Meanwhile, \textbf{Hyena} \cite{poli2023hyena} adopted a similar architecture to H3 but replaced the S4 layer with a global convolution parameterized by a multilayer perceptron (MLP) \cite{romero2021ckconv}. Building upon this groundwork, \cite{mamba} introduced a novel class of models known as \textbf{selective structured state space sequence models (S6)}, positioned as a competitive alternative to transformers in sequence modeling. This development spurred our interest in integrating \textbf{Mamba} into the skeleton action recognition landscape, which necessitates substantial temporal modeling.

\section{Methodology}
\label{sec: method}
In this section, we get acquainted with the terminology which will
be used throughout the paper and also elaborate on each component of our proposed
Simba module in more detail.

\subsection{Down-sampling ShiftGCN Encoder}
The Down-sampling ShiftGCN Encoder comprises a series of three Shift S-GCN \cite{cheng2020shiftgcn} blocks. Preceding its utilization, the input tensor undergoes an initial Shift S-GCN operation, enlarging the channel dimension of the tensor destined for the ShiftGCN encoder. This augmentation facilitates subsequent down-sampling by a factor of 2 within each Shift S-GCN of the encoder (except the last one). The central objective of the ShiftGCN encoder lies in the extraction of intricate spatial details while concurrently reducing the channel dimension of the node embeddings within the skeleton. This strategy enables a balanced trade-off between accuracy and computational efficiency as the output of this stage will be fed to the Mamba block wherein an optimal embedding dimension ($D^p \in \mathbb{R}^{V \times D}$) is essential. This dimensionality ensures effective information encapsulation without compromising computational efficacy. Mathematically, this stage can be represented by a function $ShiftGCN_{down}(x^l)$, where $x^l$ is the tensor obtained after passing through the initial Shift S-GCN, before it enters the current block :
\begin{gather}
x^l_2 = \textit{ShiftSGCN}(x^l) \\
x^l_3 = \textit{ShiftSGCN}(x^l_2) \\
x^l_4 = \textit{ShiftSGCN}(x^l_3) 
\end{gather}
At this point  $x^l_4 \in \mathbb{R}^{N \times D \times T \times V}$, where $N$ is the effective batch size, $D$ is the channel dimension, $T$ is the temporal dimension and $V$ corresponds to the number of vertices in the graph. We permute and flatten this tensor to get an output of shape $\mathbb{R}^{N \times T \times (V*D)}.$ This is subsequently fed into the intermediate mamba block.

\begin{figure}[tb]
  \centering
  \includegraphics[width=12cm,height=4.5cm]{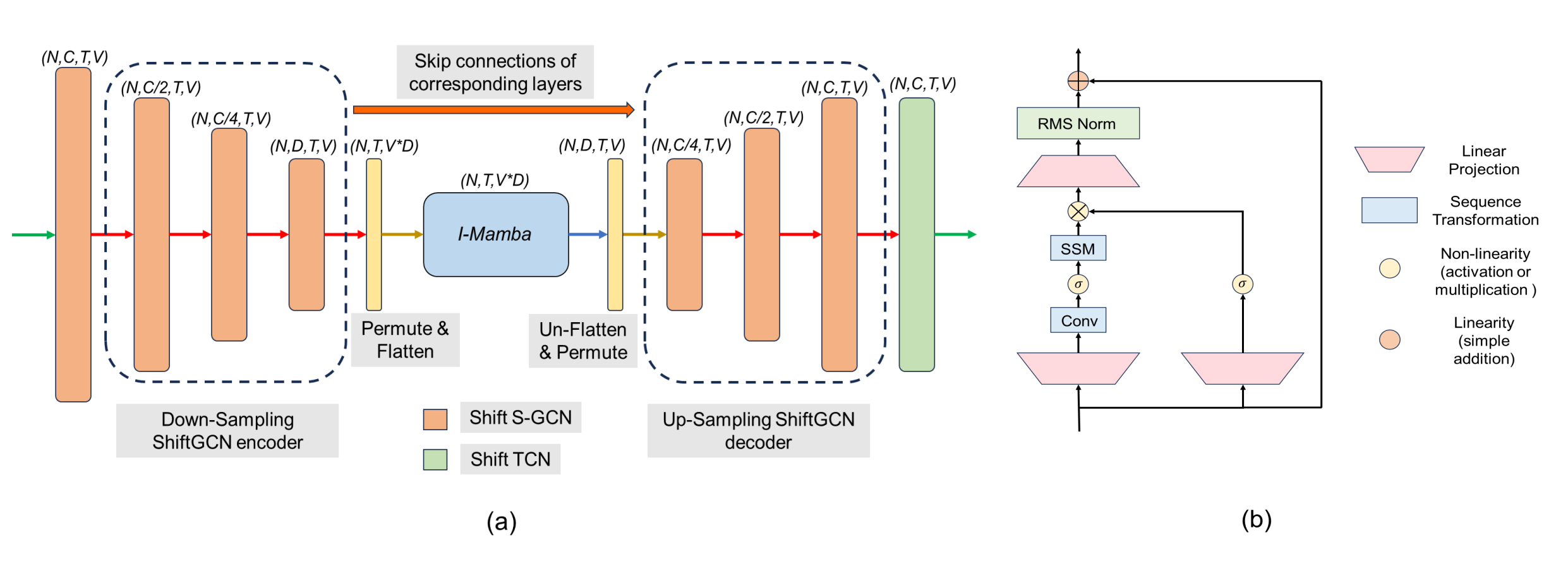}
  \caption{(a) The constituent module of our proposed model: Simba. It is composed of 4 stages or key parts: Down-sampling Shift S-GCN encoder, Intermediate Mamba block, Up-sampling Shift S-GCN decoder and a final Shift T-GCN (ShiftTCN) to enhance the temporal representation. We stack this module serially to obtain our model Simba. Each of these components is explained in details in Sec. \ref{sec: method}. The dimensions of the output tensor of each block is written at the top of their respective block.
  (b) Intermediate-Mamba (I-Mamba) block. The SSM, here, is primarily responsible for efficiently modeling long sequences like pose snapshots of videos over a given window size. This block lies at the heart of our architecture and its functionality is elaborated in subsection \ref{subsec:mamba}.
  }
  \label{fig:block_model}
\end{figure}

\subsection{Intermediate Mamba Block}
\label{subsec:mamba}

\textbf{SSM Fundamentals}. The structured state space sequence models (S4) and Mamba, both based on state-space modeling (SSM), draw inspiration from continuous systems. These models function by translating a one-dimensional function or sequence \( x(t) \) from the real number space \( \mathbb{R} \) to another real number space \( y(t) \). This translation is facilitated by an underlying hidden state \( h(t) \), residing in \( \mathbb{R}^W \). Central to this system are parameters \( A \), \( B \), and \( C \), where \( A \) governs the evolution of the hidden state \( h(t) \) and \( B \) and \( C \) act as projection parameters. Specifically, \( A \) resides in \( \mathbb{R}^{W \times W} \), while \( B \) and \( C \) belong to \( \mathbb{R}^{W \times 1} \) and \( \mathbb{R}^{1 \times W} \) respectively.

\begin{equation}
    h'(t) = Ah(t) + Bx(t), \quad y(t) = Ch(t).
    \label{eq:ssm}
\end{equation}

The S4 and Mamba represent discretized counterparts of the continuous system, incorporating a temporal scaling parameter \( \Delta \) to convert the continuous parameters \( A \) and \( B \) into their discrete counterparts \( \overline{\textit{A}} \) and \( \overline{\textit{B}} \). The prevailing technique for this conversion is the zero-order hold (ZOH) method, delineated as follows:

\begin{equation}
     \overline{\textit{A}}  = \exp(\Delta A), \quad \overline{B} = (\Delta A)^{-1} (\exp(\Delta A) - I) \cdot \Delta B.
\end{equation}

Upon discretizing \( \overline{A} \) and \( \overline{B} \), the discretized form of Eq. \ref{eq:ssm} with a step size \( \Delta \) is reformulated as:

\begin{equation}
h_t = \overline{A} h_{t-1} + \overline{B} x_t, \quad y_t = C h_t. 
\end{equation}
Finally, the models generate output via a global convolution operation.
\begin{equation}
\overline{K} = (C\overline{B}, C\overline{AB}, \dots, C\overline{A}^{M-1}\overline{B}),
\end{equation}
where \( M \) denotes the length of the input sequence \( x \), and \( \overline{K} \in \mathbb{R}^M \) represents a structured convolutional kernel.

Diverging from the predominant focus on linear time-invariant (LTI) state-space models (SSMs), Mamba distinguishes itself by integrating the selective scan mechanism (S6) \cite{mamba} as its core SSM operator. Within S6, the matrices $B \in \mathbb{R}^{B \times L \times W}$, $C \in \mathbb{R}^{B \times L \times W}$, and $\Delta \in \mathbb{R}^{B \times L \times D}$ are derived from the input data $x \in \mathbb{R}^{B \times L \times D}$. This suggests that S6 is aware of the contextual information embedded in the input, thereby ensuring the dynamic modulation of weights within this mechanism. In essence, the model transitions from being time-invariant to becoming time-varying.

\textbf{Block Equations}. We couple the Mamba \cite{mamba} block as shown in Fig. \ref{fig:block_model} (b) with RMS Normalization and a residual connection to form the intermediate mamba block. This block takes its input as a tensor of shape \( \mathbb{R}^{N \times T \times D^p} \)(where $D^p = {V \times D}$) and gives an output of the same shape. Within this block, rigorous temporal relational modeling of pose snapshots ensure that the output of this stage is rich in temporal information and the latent state of this encoder-decoder architecture is of enhanced quality.
Mathematically, the operation of the intermediate mamba block can be denoted by $\textit{IMamba}(x^l_4)$ , where $x^l_4$ is the output from the down-sampling ShiftGCN encoder stage:

\begin{gather}
B = f_B(x) , \quad C = f_C(x) , \quad \Delta = \tau_{\Delta}(P + s_{\Delta}(x))  \\
y = \textit{$\sigma$(Conv1d(Linear}(x^l_4))), \quad z = \textit{$\sigma$(Linear}(x^l_4)) \\
x^m = \textit{RMSNorm(Linear(SSM}(A,B,C,y)*z)) + x^l_4
\end{gather}

where $\sigma$ is the \textit{Silu(.)} activation, * corresponds to simple multiplication between matrices, P is $\Delta$'s parameter and A is a parameter matrix belonging to $\mathbb{R}^{D^p \times W}$. We specifically choose $f_B(x) = \text{Linear}^{ W} (x)$, $f_C(x) = \textit{Linear}^W (x)$, 
$s_{\Delta}(x) = \textit{Broadcast}^{D^p}(\textit{Linear}^1(x))$, and $\tau_{\Delta} = \textit{softplus}$, where $\textit{Linear}^d$ is a parameterized projection to dimension $d$ (\textit{Broadcast} holds the same notation).

We unflatten the last dimension of $x^m$ and permute it to make it a tensor of shape $\mathbb{R}^{N \times D \times T \times V}$, which also makes it compatible with the decoder section of the Simba architecture.

\subsection{Up-sampling ShiftGCN Decoder}
Similar to the down-sampling segment, the up-sampling pathway within Simba module comprises three consecutive Shift S-GCN blocks. These blocks increment the channel dimension by a factor of 2 per block, with the exception of the initial Shift S-GCN, which translates the dimension D to C. Essentially, the up-sampling pathway serves a dual purpose: firstly, to facilitate an increase in the channel dimension, thereby alleviating information loss during the encoding phase, and secondly, to ensure compatibility of the output with subsequent Simba modules. Formally, the up-sampling ShiftGCN decoder can be represented by $ShiftGCN_{up}(x^n)$, where $x^n$ is the un-flattened and permuted version of $x^m$ as mentioned in subsection \ref{subsec:mamba} :
\begin{gather}
x^l_5 = \textit{ShiftSGCN}(x^n) + x^l_3\\
x^l_6 = \textit{ShiftSGCN}(x^l_5) + x^l_2\\
x^l_7 = \textit{ShiftSGCN}(x^l_6) + x^l
\end{gather}

The skip connections adhere to the U-Net framework as outlined by Ronneberger et al. \cite{ronneberger2015u}, serving to prevent the occurrence of vanishing gradients within the model. Additionally, these connections serve the crucial function of preserving the initial information provided to the model, even after undergoing several convolutional operations across its depth. The output of the up-sampling ShiftGCN decoder undergoes a final refinement process facilitated by a Shift T-GCN \cite{cheng2020shiftgcn}, also referred to as ShiftTCN. This additional step aims to augment the temporal dimension further, thereby enhancing the overall quality of the output.

\subsection{Overall Model Architecture}
Assume that the input to the $l^{th}$ layer of Simba is $x(l)$, then the overall architecture of the Simba module can be summed up using the following equations:
\begin{gather}
x^l = \textit{ShiftSGCN}(x(l)) \\
x^l_4 = ShiftGCN_{down}(x^l) \\
x^l_4 = \textit{Flatten(Permute}(x^l_4)) \\
x^m = \textit{IMamba}(x^l_4) \\
x^n = \textit{Permute(UnFlatten}(x^m)) \\
x^l_7 = ShiftGCN_{up}(x^n) \\
x(l+1) = \rho(\textit{ShiftTCN}(x^l_7) + \textit{Residual}(x(l))),
\end{gather}
where $x(l+1)$ is the input to the $(l+1)^{th}$ layer of Simba, $\rho$ is the \textit{ReLU(.)} actiavtion function and \textit{Residual(.)} is simply a unit TCN composed of a 2-D convolutional layer followed by a 2-D Batch Normalization layer (akin to the implementation of \cite{cheng2020shiftgcn,cheng2021extremely}).

We transmit the output from the final Simba module through a Fully Connected layer to transform the tensor into the dimension corresponding to the number of classes within the datasets. Consistent with prior studies such as \cite{cheng2020shiftgcn,cheng2021extremely,shi2020decoupled}, we employ a Cross Entropy loss function for training our model. Further elaboration is provided in subection \ref{subsec:implementation_det}.

\textbf{Intuition Build-up.} The encoder-decoder architecture of the Simba model may suggest to use simpler layers to down-sample and up-sample the channel dimension of the input tensor. However, we have observed that using linear layers to accomplish this task becomes detrimental to the performance of the network. What is needed is a gradual shrinkage of the channel space and a smooth enlargement of the same space. Intuitively, this can be accomplished by a U-Net \cite{ronneberger2015u} architecture where the backbone should be capable of squeezing out fruitful details just like our Shift-GCN backbone. It is therefore, of paramount importance to carefully trade between the loss of information and the complexity of the Simba module.

\section{Experiments \& Results}
\label{sec:exp}
In this section, we commence by contrasting our Simba architecture with the current leading methodologies on benchmarks for skeleton-based human action recognition and demonstrate the superior efficacy of our model. Subsequently, we undertake an ablation study to delve deeper into our proposed approach for a more comprehensive understanding.

\subsection{Datasets}
We assess the efficacy of our proposed Simba on three widely recognised public datasets: NTU-RGB+D \cite{7780484}, NTU-RGB+D 120 \cite{8713892}, and Northwestern-UCLA \cite{wang2014cross}, which are briefly described as follows:


\textbf{NTU-RGB+D} \cite{7780484} dataset serves as a prominent benchmark for skeleton-based human action recognition. It comprises 56,880 skeletal action sequences, each performed by either one or two individuals. These sequences are captured by three Microsoft Kinect-V2 depth sensors, placed at the same height but from varying horizontal perspectives simultaneously. Evaluation is carried out using two distinct benchmarks: Cross-Subject (X-Sub) and Cross-View (X-View) settings. In the X-Sub setup, training and test datasets are drawn from two separate cohorts of 20 subjects each. In the X-View scenario, the training set consists of 37,920 samples captured by camera views 2 and 3, while the test set comprises 18,960 sequences recorded by camera view 1.

\textbf{NTU-RGB+D 120} \cite{8713892} dataset extends the NTU-RGB+D dataset by including an additional 57,367 skeleton sequences across 60 supplementary action classes. It currently stands as the largest available dataset with 3D joint annotations for human action recognition, featuring 32 setups, each representing distinct locations and backgrounds. Two benchmark evaluations were suggested by the author including Cross-Subject (X-Sub) and Cross-Setup (X-Setup) settings.

\textbf{Northwestern-UCLA} \cite{wang2014cross} dataset is captured by three Kinect sensors from various viewpoints and comprises 1,494 video sequences across 10 action categories.

\subsection{Implementation details}
\label{subsec:implementation_det}
\textbf{NTU RGB+D 60} and \textbf{NTU RGB+D 120.} Our models, Simba and U-ShiftGCN, are trained for 90 epochs, achieving convergence sooner than the typical 140 epochs employed in previous studies \cite{cheng2020shiftgcn,cheng2021extremely}. The learning rate begins at 0.025 and experiences decay of 0.1 at epochs 75 and 85. Training and testing batches consist of 64 and 512 sizes, respectively. The window size (\textit{T}), representing the sampled number of frames per video, is set to 64 during data pre-processing following the approach outlined in \cite{zhang2020semantics}.

\textbf{Northwestern-UCLA.} The training and testing batch sizes are set to 16 and 64, respectively. A window size of 52 is utilized. Our models are trained for 400 epochs to maximize their potential and we also adhere to the pre-processing strategy outlined in \cite{zhang2020semantics}.

We apply a weight decay of 0.0001, consistent with previous works \cite{cheng2020shiftgcn,cheng2021extremely}, for both NTU RGB+D 60 and NTU RGB+D 120 datasets, while for NW-UCLA, the weight decay is set to 0.0004. In configuring the mamba block, we fix the embedding dimension (d-model) at 500, aligning closely with the base mamba architecture \cite{mamba} inspired by ViT-B \cite{dosovitskiy2020image}, where the embedding dimension is 768. To ensure consistency, we adjust the channel dimension of the mamba-adjacent Shift-GCN blocks to 20 for NTU datasets and 25 for NW-UCLA datasets, given their respective skeleton node counts of 25 and 20 respectively. We also employ $(l=10)$ as the depth of our model following \cite{shi2020decoupled,cheng2020shiftgcn,cheng2021extremely}. Further details on the implementation can be found in the supplementary materials section.

\subsection{Comparison with state-of-the-art}
In line with recent advancements in the field \cite{cheng2020shiftgcn,ye2020dynamic,shi2020decoupled}, we embrace a multi-stream fusion approach. Specifically, we incorporate four streams, each tailored to a distinct modality: \textbf{joint, bone, joint motion,} and \textbf{bone motion}. The joint modality encompasses the raw skeleton coordinates, while the bone modality captures spatial coordinate differentials. On the other hand, the joint motion and bone motion modalities focus on temporal differentials within the joint and bone modalities, respectively. To consolidate the information from these streams, we aggregate the softmax scores from each stream to derive the fused score.

\begin{table}[]
\centering
\caption{Results for NW-UCLA.\textsuperscript{\dag}Re-trained from scratch using the official code on our GPU system.}
\label{tab:UCLA}
\begin{tabular}{@{}ccc@{}}
\toprule
Methods           & Year & Top-1  \\ \midrule
Lie Group\cite{veeriah2015differential}         & 2015 & 74.2   \\
HBRNN-L\cite{Du_2015_CVPR}           & 2015 & 78.5   \\
Ensemble TS-LSTM\cite{Lee_2017_ICCV}  & 2017 & 89.2   \\
2s AGC-LSTM\cite{8954298}       & 2019 & 93.3   \\
VA-CNN(aug.)\cite{zhang2019view}      & 2019 & 90.7   \\
1s Shift-GCN\cite{cheng2020shiftgcn}      & 2020 & $89.85^\dag$ \\
4s Shift-GCN\cite{cheng2020shiftgcn}      & 2020 & 94.6   \\
DC-GCN +ADG\cite{shi2020decoupled}       & 2020 & 95.3   \\
1s Shift-GCN++\cite{cheng2021extremely}      & 2021 & 93.8 \\
4s Shift-GCN++\cite{cheng2021extremely}      & 2021 & 95.0   \\
Ta-CNN\cite{xu2022topology}            & 2022 & 96.1   \\
GAP(joint)\cite{xiang2022language}        & 2023 & 94.0   \\ \midrule
U-ShiftGCN(joint) & Ours & 91.81  \\ 
Simba(joint)      & Ours & 94.18  \\ 
Simba(4-ensemble) & Ours & {96.34} \\ \bottomrule
\end{tabular}
\end{table}

\begin{table}[]
\centering
\caption{Results for NTU RGB+D 60. \textsuperscript{\dag}Shift-GCN was retrained using our GPU and the original settings, yielding results less favorable than those reported in the original paper. \textsuperscript{\ddag}Additionally, AutoGCN achieved a 95\% confidence interval of [86.9, 89.2] for X-Sub and [94.6, 96.3] for X-View, as reported in their paper.}
\label{tab:NTU60}
\begin{tabular}{@{}cclcc@{}}
\toprule
\multirow{2}{*}{Methods} & \multicolumn{2}{c}{\multirow{2}{*}{Year}} & \multirow{2}{*}{X-Sub(\%)} & \multirow{2}{*}{X-View(\%)} \\
                         & \multicolumn{2}{c}{}                      &                            &                             \\ \midrule
VA-LSTM\cite{zhang2017view}                  & \multicolumn{2}{c}{2017}                  & 79.4                       & 87.6                        \\
ST-GCN\cite{yan2018spatial}                & \multicolumn{2}{c}{2018}                  & 81.5                       & 88.3                        \\
SR-TSL\cite{si2018skeleton}                   & \multicolumn{2}{c}{2018}                  & 84.8                       & 92.4                        \\
Motif  + VTDB\cite{Wen_Gao_Fu_Zhang_Xia_2019}            & \multicolumn{2}{c}{2019}                  & 84.2                       & 90.2                        \\
2s AS-GCN\cite{Li_2019_CVPR}                & \multicolumn{2}{c}{2019}                  & 86.8                       & 94.2                        \\
1s-AGCN\cite{li2018adaptive}                & \multicolumn{2}{c}{2019}                  & 86.0                       & 93.7                        \\
1s Shift-GCN\cite{cheng2020shiftgcn}             & \multicolumn{2}{c}{2020}                  & 87.8                       & $93.0^{\dag}$                       \\
TS-SAN\cite{cho2020self}                   & \multicolumn{2}{c}{2020}                  & 87.2                       & 92.7                        \\
MS-TGN\cite{li2020multi}                   & \multicolumn{2}{c}{2020}                  & 86.6                       & 94.1                        \\
MS TE-GCN\cite{li2020temporal}                & \multicolumn{2}{c}{2020}                  & 87.4                       & 93.4                        \\
NAS-GCN(joint)\cite{Peng_Hong_Chen_Zhao_2020}              & \multicolumn{2}{c}{2020}                  & 87.5                       & 94.7                        \\
1s Shift-GCN++\cite{cheng2021extremely}           & \multicolumn{2}{c}{2021}                  & 87.9                       & 94.8                        \\
1s IIP-Transformer\cite{10386970}       & \multicolumn{2}{c}{2023}                  & 88.9                       & 94.2                        \\
SNAS-GCN(joint)\cite{electronics12143156}                 & \multicolumn{2}{c}{2023}                  & 87.1                       & 94.3                        \\
AutoGCN\cite{tempel2024autogcn}                  & \multicolumn{2}{c}{2024}                  & $88.3^\ddag$                       & $95.5^\ddag$                        \\ \midrule
Simba(joint)       & \multicolumn{2}{c}{Ours}                  & 89.03                      & 94.38                       \\ \bottomrule
\end{tabular}
\end{table}

\begin{table}[]
\centering
\caption{Results for NTU RGB+D 120. }
\label{tab:NTU120}
\begin{tabular}{@{}cclcc@{}}
\toprule
\multirow{2}{*}{Methods} & \multicolumn{2}{c}{\multirow{2}{*}{Year}} & \multirow{2}{*}{X-Sub(\%)} & \multirow{2}{*}{X-Set(\%)} \\
                         & \multicolumn{2}{c}{}                      &                            &                             \\ \midrule
ST-GCN\cite{yan2018spatial}                & \multicolumn{2}{c}{2018}                  & 70.7                      & 73.2                        \\
AS-GCN\cite{Li_2019_CVPR}                & \multicolumn{2}{c}{2019}                  & 77.9                       & 78.5                        \\
2s-AGCN\cite{li2018adaptive}                & \multicolumn{2}{c}{2019}                  & 82.5                       & 84.2                        \\
1s Shift-GCN\cite{cheng2020shiftgcn}             & \multicolumn{2}{c}{2020}                  & 80.9                       & 83.2                       \\
SGN\cite{zhang2020semantics}             & \multicolumn{2}{c}{2020}                  & 79.2                     & 81.5                       \\
MS-G3D(joint)\cite{liu2020disentangling}             & \multicolumn{2}{c}{2020}                  & 82.3                     & 84.1                       \\
ST-TR\cite{plizzari2021spatial}             & \multicolumn{2}{c}{2021}                  & 82.7                     & 84.7                       \\
AutoGCN\cite{tempel2024autogcn}                  & \multicolumn{2}{c}{2024}                  & 83.3                      & 84.1                        \\ \midrule
Simba(joint)       & \multicolumn{2}{c}{Ours}                  & 79.75                      & 86.28                       \\ \bottomrule
\end{tabular}
\end{table}

The comparison on the three datasets is shown in Tables \ref{tab:UCLA},\ref{tab:NTU60} and \ref{tab:NTU120}, respectively. As depicted in Table \ref{tab:NTU60}, our model achieves performance better or atleast comparable with the state-of-the-art in both the settings: Cross-subject and Cross-view for NTU RGB+D 60 dataset. The Northwestern-UCLA dataset poses a notable challenge due to its limited training samples, thereby intensifying the difficulty for models to excel. Despite this demanding scenario, our Simba model demonstrates its remarkable capability in efficiently capturing intermediate temporal dynamics, leading to state-of-the-art performance as shown in Table \ref{tab:UCLA}. As shown in Table \ref{tab:NTU120}, for NTU RGB+D 120, our model Simba performs exceptionally well, exceeding the state-of-the-art on the x-set setting and achieving more than 86\%.

\subsection{Ablation study}

\textbf{Effect of IMamba.} To assess the impact of the Intermediate Mamba Block at the core of our Simba architecture, we simply remove the IMamba block (since the IMamba block expects input and output tensors of the same shape). This ablated model is what we have dubbed as U-ShiftGCN. As depicted in Table \ref{tab:ablation1}, our novel mamba augmented Simba model (joint only modality) achieves 94.18\% on NW-UCLA dataset marking a substantial increase of 2.37\% in comparison with U-ShiftGCN (joint only modality) and an enhancement of 4.33\% over our baseline model, 1s Shift-GCN. This underscores the principal contribution of our work, highlighting the significance of mamba integration for intermediate temporal modeling in skeletal action recognition (SAR).
\begin{table}[]
\centering
\caption{Ablation on NW-UCLA}
\label{tab:ablation1}
\begin{tabular}{@{}cc@{}}
\toprule
Model             & Accuracy(\%) \\ \midrule
1s Shift-GCN      & 89.85        \\
U-ShiftGCN(joint) & 91.81        \\ 
Simba(joint)      & 94.18       \\ \bottomrule
\end{tabular}
\end{table}

\textbf{Optimal number of layers.} Empirally, we experimented with the number of layers for our U-ShiftGCN architecture. As shown in Table \ref{tab:ablation2}, 10 layers achieves the best performance among the set \{6,10,12\} for number of layers. Simba also follows this trend akin to previous architectures \cite{cheng2020shiftgcn,shi2020decoupled,cheng2021extremely}. Hence, this justifies our choice of using $(l=10)$ as the depth for this model.
\begin{table}[]
\centering
\caption{Effect of number of layers on U-ShiftGCN}
\label{tab:ablation2}
\begin{tabular}{@{}cc@{}}
\toprule
Number of layers & Accuracy(\%) \\ \midrule
6                & 86.75        \\
10               & 91.81        \\ 
12               & 88.16       \\ \bottomrule
\end{tabular}
\end{table}

\section{Conclusion}
\label{sec: conclusion}
To encapsulate, our work pioneers the integration of Mamba, a selective state space model, into the domain of Skeleton Action Recognition (SAR) and graph data. By leveraging Mamba within a novel encoder-decoder architecture with a Shift-GCN backbone, we address the challenge of efficiently modeling long sequences inherent in SAR tasks. Unlike plain Transformers, which lack structural priors and underperform compared to leading GCN-based methods, our approach harnesses the power of Mamba to enhance temporal modeling while preserving spatial information.
Our proposed model, dubbed Simba, achieves state-of-the-art performance across three benchmark SAR datasets: NTU RGB+D, NTU RGB+D 120, and Northwestern-UCLA. Notably, even without the Intermediate Mamba Block, the U-ShiftGCN component of Simba surpasses baseline performance, highlighting the efficacy of our approach.
The essence of our success lies in the careful design of the encoder-decoder architecture, which incorporates downsampling and upsampling spatial and temporal subunits while ensuring a smooth transition in the channel space. Unlike using linear layers, which can detrimentally affect network performance, our approach achieves a gradual shrinkage and enlargement of the channel space, akin to the dynamics of a U-Net architecture. The meticulous balance struck between mitigating information loss and managing complexity epitomizes the innovative and inventive nature of our methodology. This delicate equilibrium not only underscores the novelty of our approach but also highlights its creativity in addressing the intricacies of Skeleton Action Recognition (SAR) tasks. By seamlessly integrating Mamba into our framework, we unlock a wealth of opportunities for further advancement in SAR research.
The incorporation of Mamba not only yields significant benefits for SAR tasks but also paves the way for the integration of other encoder-decoder architectures within our framework. With Mamba at the core of these architectures, we are poised to tackle the challenges inherent in SAR with renewed vigor and effectiveness. This forward-looking approach holds promise for pushing the boundaries of SAR performance and advancing the state-of-the-art in the field.

\clearpage

\appendix 

\section*{Supplementary Materials}

\section{Partition Gating Mechanism}
In Simba architecture, for NTU RGB+D 60 and NTU RGB+D 120 datasets, we adopt a particular partition gating mechanism that is seen to enhance model performance. Suppose \textit{W} is a learnable parameter matrix belonging to $\mathbb{R}^{1 \times C \times 1 \times 1}$, where \textit{C} is the input channel dimension. Let \textit{K} be the number of partitions and \textit{V} the number of vertices of the skeletal graph. We define \textit{label} as the one-hot encoded vector of the joint partition-membership labels having shape $\mathbb{R}^{V \times \textit{K}}$. Here, $x^l$ denotes the output of the initial Shift S-GCN block in the Simba module, \textit{proj(.)} denotes a 2-D convolution operation that transforms channel dimension $C$ to $C'$ and $\Pi$ is a selection operation that fetches for every joint node, the partition tensor belonging to $\mathbb{R}^{N \times C' \times T}$, responsible for describing it among the \textit{K} possible representations for the partitions denoted by the normalized tensor \textit{z}:

\begin{gather}
    \textit{label}_{ij} = \begin{cases} 1 & \text{if } \textit{joint(i)} \in partition(j) \\ 0 & \text{otherwise} \end{cases}, \quad i \in [1, V], \ j \in [1, \textit{K}] \\
    z = x(l) * \left(\frac{\textit{label}}{\sum_{i=1}^{V} \textit{label}_{i}}\right), \quad z \in \mathbb{R}^{N \times C \times T \times \textit{K}} \\
    e = \Pi(proj(z)), \quad e \in \mathbb{R}^{N \times C' \times T \times \textit{V}}
\end{gather}
Then, we define the partition augmented input, $x^l_p$, using a gating mechanism as:
\begin{gather}
    x^l_p = W*x(l) + (1-W)*e
\end{gather}
where * is the matrix multiplication. We pass this partition enhanced input, $x^l_p$, to $ShiftGCN_{down}(.)$ instead of $x^l$. This valuable partition incorporation is highly beneficial for large datasets like NTU, where the model is able to learn from both joint-level as well as group or partition level information simultaneously leading to better performance. 

\section{Extended NTU RGB+D 60 results}
We present an extended and more comprehensive version of our NTU RGB+D 60 results in Table \ref{tab:NTU60}. We have compared the results of Simba 2-ensemble (joint + bone) with the state-of-the-art methodologies.

\begin{table}[htbp]
\centering
\caption{Extended version of the results for NTU RGB+D 60. \textsuperscript{\dag}Shift-GCN was retrained using our GPU and the original settings, yielding results less favorable than those reported in the original paper. \textsuperscript{\ddag}Additionally, AutoGCN achieved a 95\% confidence interval of [86.9, 89.2] for X-Sub and [94.6, 96.3] for X-View, as reported in their paper.}
\label{tab:NTU60}
\begin{tabular}{@{} c @{\hspace{5mm}} c @{\hspace{5mm}} c @{\hspace{5mm}} c @{\hspace{5mm}} c @{\hspace{5mm}} c @{\hspace{5mm}} c @{\hspace{5mm}} c @{}}
\toprule
\multirow{2}{*}{Methods} & \multirow{2}{*}{Year} & \multicolumn{3}{c}{X-Sub} & \multicolumn{3}{c}{X-View} \\  
                         &                       & \textit{J}       & \textit{B}      & \textit{J+B}    & \textit{J}       & \textit{B}       & \textit{J+B}    \\ \cmidrule(r){1-8}
VA-LSTM \cite{zhang2017view}                   & 2017                  &79.40    & -   &-   & 87.60    & -    & -   \\
ST-GCN \cite{yan2018spatial}                    &2018                       &81.50         &-        &-        &88.30         &-       &-      \\
SR-TSL \cite{si2018skeleton}                   & 2018                  & 84.80    & -   & -   & 92.40    & -    & -   \\
Motif+VTDB \cite{Wen_Gao_Fu_Zhang_Xia_2019}                   & 2019                  & 84.20    & -   & -   & 90.20    & -    & -   \\
2s-AGCN \cite{li2018adaptive}                    &2019                       &-         &-        &88.5        &93.70         &  93.20       &  95.10      \\
TS-SAN\cite{cho2020self}                         &2020                       &87.20         &-        &-        &92.70         &-         &-        \\
MS-TGN\cite{li2020multi}                         &2020                       &86.60         &87.50        &89.50        &94.10         &93.90         &95.90        \\
3s RA-GCN\cite{song2020richly}                         &2020                       &-         &-        &87.30        &-         &-         &93.60        \\
2s Shift-GCN\cite{cheng2020shiftgcn}                         &2020                       &-         &-        &$88.50^{\dag}$        &-         &-         &$94.10^{\dag}$        \\
NAS-GCN\cite{Peng_Hong_Chen_Zhao_2020}                         &2020                       &87.50         &-        &89.40        &94.70         &-         &95.70        \\
UNIK\cite{yang2021unik}                         &2021                       &-         &-        &86.80        &-         &-         &94.40        \\
AdaSGN\cite{shi2021adasgn}                         &2021                       &-         &-        &89.10        &-        &-         &94.70        \\
SNAS-GCN\cite{li2020multi}                         &2023                       &87.10       &-        &89.00        &94.30         &-         &95.00        \\
AutoGCN\cite{tempel2024autogcn}                         &2024                       &$88.30^{\ddag}$       &-        &-        &$95.50^{\ddag}$        &-         &-        \\ \midrule
Simba                         &Ours                       &89.03       &88.48        &90.54        &94.38        &93.41         &95.21        \\\bottomrule
\end{tabular}
\end{table}

\section{Extended Training Recipe}
In Table \ref{tab:config}, we have made available the configuration settings that we have used in our experiments for the three benchmark datasets.

\begin{table}[]
\centering
\caption{Major configuration settings for the three datasets that we have used in our experiments}
\label{tab:config}
\begin{tabular}{|c|ccc|}
\hline
\textbf{Configuration}          & \multicolumn{1}{c|}{\textbf{NW-UCLA}}   & \multicolumn{1}{c|}{\textbf{NTU RGB+D 60}} & \textbf{NTU RGB+D 120} \\ \hline
Training Batch Size    & \multicolumn{1}{c|}{16}        & \multicolumn{2}{c|}{64}                           \\ \hline
Test Batch Size        & \multicolumn{1}{c|}{64}        & \multicolumn{2}{c|}{512}                          \\ \hline
Weight Decay           & \multicolumn{1}{c|}{0.0004}    & \multicolumn{2}{c|}{0.0001}                       \\ \hline
Window Size            & \multicolumn{1}{c|}{52}        & \multicolumn{2}{c|}{64}                           \\ \hline
Number of Epochs       & \multicolumn{1}{c|}{400}       & \multicolumn{2}{c|}{90}                           \\ \hline
Step                   & \multicolumn{1}{c|}{{[}110{]}} & \multicolumn{2}{c|}{{[}75,85{]}}                  \\ \hline
Repeat Augmentation    & \multicolumn{1}{c|}{True}      & \multicolumn{2}{c|}{False}                        \\ \hline
Optimizer              & \multicolumn{3}{c|}{SGD}                                                           \\ \hline
Base Lr                & \multicolumn{3}{c|}{0.025}                                                         \\ \hline
Lr decay rate          & \multicolumn{3}{c|}{0.1}                                                           \\ \hline
Warm-up epochs         & \multicolumn{3}{c|}{5}                                                             \\ \hline
Channel dimension(C)   & \multicolumn{3}{c|}{216}                                                           \\ \hline
Mamba hidden dimension & \multicolumn{3}{c|}{500}                                                           \\ \hline
\end{tabular}
\end{table}

\section{Pseudo-code for Simba}
The following is the pseudo-code for our novel Simba module without partition gating mechanism (Replace $x^l$ with $x^l_p$ to get the corresponding pseudo-code involving the mechanism):
\begin{algorithm}
    \SetAlgoLined
    \KwIn{graph sequence $x(l) : (N, C_i, T, V)$}
    \KwOut{graph sequence $x(l+1) : (N, C, T, V)$}
    \tcp{Map channel dimension $C_i$ to $C$}
    $x^l: (N,C,T,V) \leftarrow \text{ShiftSGCN}(x(l))$\;
    \tcp{Downsampling Shift-GCN Encoder}
    $x^l_4: (N,D,T,V) \leftarrow \text{ShiftGCN}_{down}(x^l)$\;
    \tcp{Making a single vector embedding for each graph snapshot}
    $x^l_4: (N, T, D^p) \leftarrow \text{Flatten(Permute}(x^l_4))$\;
    \tcp{Intermediate Temporal Modeling via I-Mamba}
    $x^m: (N, T, D^p) \leftarrow \text{IMamba}(x^l_4)$\;
    $x^n: (N, D, T, V) \leftarrow \text{Permute(UnFlatten}(x^m))$\;
    \tcp{Upsampling Shift-GCN Decoder}
    $x^l_7: (N, C, T, V) \leftarrow \text{ShiftGCN}_{up}(x^n)$\;
    $x(l+1): (N, C, T, V) \leftarrow \text{Relu}(\text{ShiftTCN}(x^l_7)+ \text{Residual}(x(l)))$\;

    \KwRet{$x(l+1)$}
    \caption{Simba Module process}
\end{algorithm}



%
%
\bibliographystyle{splncs04}
\bibliography{main}
\end{document}